\documentclass[11pt,a4paper]{article}
\usepackage[hyperref]{acl2017}
\usepackage{times}
\usepackage{latexsym}
\usepackage{amsmath}
\usepackage{amssymb}
\usepackage{graphicx}
\usepackage{ifthen}
\usepackage[utf8]{inputenc}
\usepackage{url}
\usepackage{xspace}

\aclfinalcopy

\newcommand\sZ{\ensuremath{\mathcal{Z}}}

\newcommand\bu{\ensuremath{\mathbf{u}}}

\newcommand\bz{\ensuremath{\mathbf{z}}}

          \newcommand\refeqn[1]{(\ref{eqn:#1})}

\newcommand\refsec[1]{Section~\ref{sec:#1}}

\newcommand\reffig[1]{Figure~\ref{fig:#1}}

\newcommand\reftab[1]{Table~\ref{tab:#1}}
\newcommand\refapp[1]{Appendix~\ref{sec:#1}}

\ifthenelse{\isundefined{\definition}}{\newtheorem{definition}{Definition}}{}
\ifthenelse{\isundefined{\assumption}}{\newtheorem{assumption}{Assumption}}{}
\ifthenelse{\isundefined{\hypothesis}}{\newtheorem{hypothesis}{Hypothesis}}{}
\ifthenelse{\isundefined{\proposition}}{\newtheorem{proposition}{Proposition}}{}
\ifthenelse{\isundefined{\theorem}}{\newtheorem{theorem}{Theorem}}{}
\ifthenelse{\isundefined{\lemma}}{\newtheorem{lemma}{Lemma}}{}
\ifthenelse{\isundefined{\corollary}}{\newtheorem{corollary}{Corollary}}{}
\ifthenelse{\isundefined{\alg}}{\newtheorem{alg}{Algorithm}}{}
\ifthenelse{\isundefined{\example}}{\newtheorem{example}{Example}}{}

\newcommand\C[1]{\texttt{#1}}
\newcommand\JRL{J_\mathrm{RL}}
\newcommand\JMML{J_\mathrm{MML}}
\newcommand\LMML{\mathcal{L}_\mathrm{MML}}
\newcommand\policy{p_\theta(\bz \mid x)}
\newcommand\ourmodel{\textsc{RandoMer}\xspace}  \newcommand\refsubsec[1]{Subsection~\ref{subsec:#1}}
 
\title{From Language to Programs: Bridging Reinforcement Learning and Maximum Marginal Likelihood}

\author{
  Kelvin Guu \\
  Statistics \\
  Stanford University \\
  {\small \tt kguu@stanford.edu}
\And
  Panupong Pasupat \\
  Computer Science \\
  Stanford University \\
  {\small \tt ppasupat@stanford.edu}
\And
  Evan Zheran Liu \\
  Computer Science \\
  Stanford University \\
  {\small \tt evanliu@stanford.edu}
\And
  Percy Liang \\
  Computer Science \\
  Stanford University \\
  {\small \tt pliang@cs.stanford.edu}
}

\date{}

\begin{document}

\maketitle

\begin{abstract}
Our goal is to learn a semantic parser that maps natural language utterances
into executable programs when only indirect supervision is
available: examples are labeled with the correct execution result,
but not the program itself. Consequently, we must search the space of programs
for those that output the correct result, while not being misled by
\emph{spurious programs}: incorrect programs that coincidentally output the
correct result.
We connect two common learning paradigms, reinforcement learning (RL) and maximum marginal likelihood (MML),
and then present a new learning algorithm that combines the strengths of both.
The new algorithm guards against spurious programs
by combining the systematic search traditionally employed in MML with the randomized exploration of RL,
and by updating parameters such that probability is spread more evenly across consistent programs.
We apply our learning algorithm to a new neural semantic parser and
show significant gains over
existing state-of-the-art results on a recent context-dependent semantic parsing task.

 \end{abstract}

\section{Introduction}

\begin{figure}\center
\includegraphics[width=\columnwidth]{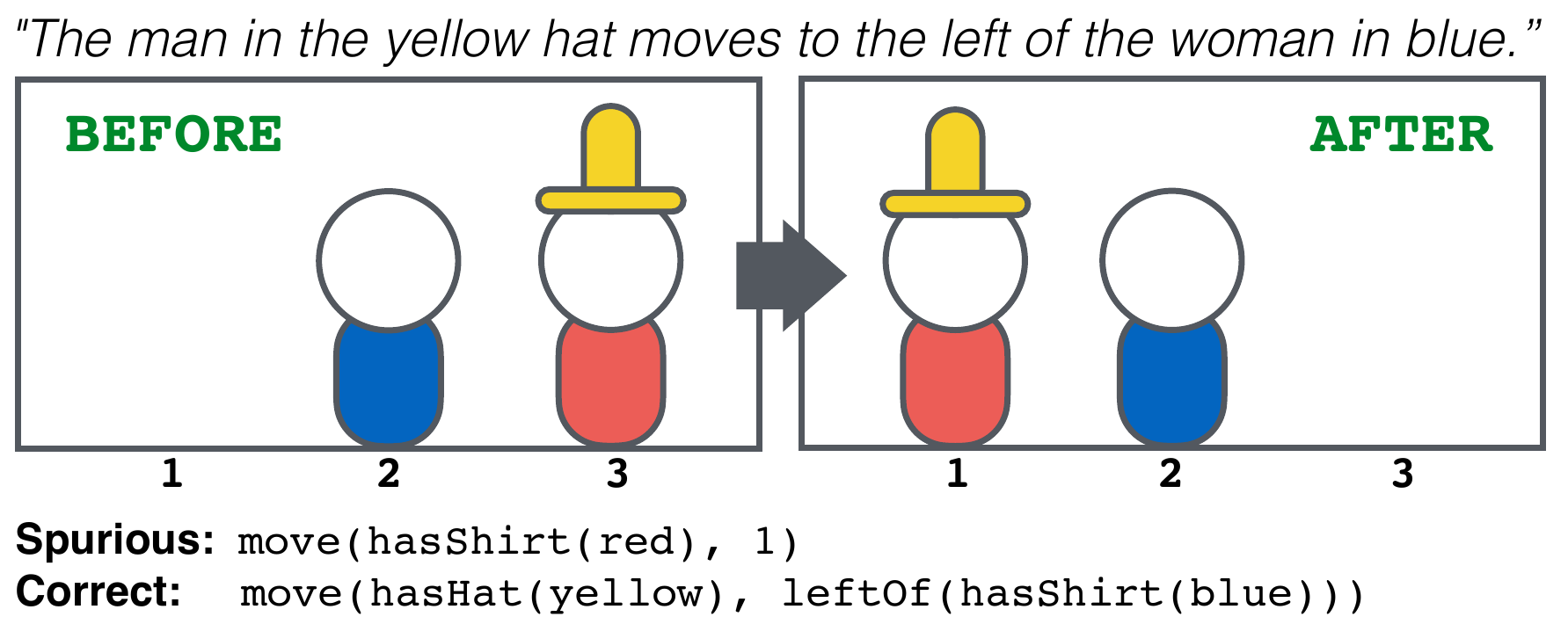}
\caption{
The task is to map natural language utterances
to a program that manipulates the world state.
The correct program captures the true meaning of the utterances,
while spurious programs arrive at the correct output
for the wrong reasons.
We develop methods to prevent the model from being drawn to spurious programs.
}\label{fig:spurious}
\end{figure}

We are interested in learning a semantic parser
that maps natural language utterances into executable programs
(e.g., logical forms).
For example, in \reffig{spurious},
a program corresponding to the utterance
transforms an initial world state into a new world state.
We would like to learn from indirect supervision, where each
training example is only labeled with the correct output (e.g. a target world
state), but not the program that produced that output \citep{clarke10world,liang11dcs,krishnamurthy2012weakly,artzi2013weakly,liang2017nsm}.

The process of constructing a program can be formulated as a
sequential decision-making process,
where feedback is only received at the end of the sequence
when the completed program is executed.
In the natural language processing literature,
there are two common approaches for handling this situation:
1) reinforcement learning (RL),
particularly the REINFORCE algorithm \citep{williams1992simple, sutton1999policy},
which maximizes the expected reward of a sequence of actions;
and
2) maximum marginal likelihood (MML), which treats the sequence of actions
as a latent variable, and then maximizes the marginal likelihood of observing
the correct program output \citep{demp1977em}. 

While the two approaches have enjoyed success on many tasks,
we found them to work poorly out of the box for our task.
This is because
in addition to the sparsity of correct programs,
our task also requires weeding out \emph{spurious programs} \citep{pasupat2016inferring}:
incorrect interpretations of the utterances that accidentally produce the correct output,
as illustrated in \reffig{spurious}.

We show that MML and RL optimize closely related objectives. Furthermore, both MML and RL methods have a
mechanism for exploring program space in search of programs that generate the
correct output. We explain why this exploration tends to quickly concentrate
around short spurious programs, causing the model to sometimes overlook the correct program. To address this problem, we propose \ourmodel, a new learning algorithm with two parts:

First, we propose \emph{randomized beam search},
an exploration strategy which combines the systematic
beam search traditionally employed in MML with the randomized off-policy
exploration of RL.
This increases the chance of finding correct programs
even when the beam size is small or the parameters are not pre-trained.

Second, we observe that even with good exploration, the gradients of both the 
RL and MML objectives may still upweight entrenched spurious programs more strongly than
correct programs with low probability under the current model.
We propose a \emph{meritocratic parameter update rule},
a modification to the MML gradient update, which more equally upweights all programs that produce the correct output.
This makes the model less likely to overfit spurious programs.

We apply \ourmodel to train a new neural semantic parser, which outputs programs
in a stack-based programming language. We evaluate our resulting system on
SCONE, the context-dependent semantic parsing dataset of
\newcite{long2016projections}. Our approach outperforms standard RL and MML
methods in a direct comparison, and achieves new state-of-the-art results,
improving over \citet{long2016projections} in all three domains of SCONE, and by
over 30\% accuracy on the most challenging one.

 \section{Task}

We consider the semantic parsing task in the SCONE dataset\footnote{
{\scriptsize\url{https://nlp.stanford.edu/projects/scone}}}
\cite{long2016projections}.
As illustrated in \reffig{spurious},
each example consists of
a \emph{world} containing several objects (e.g., people),
each with certain properties (e.g., shirt color and hat color).
Given the initial world state $w_0$
and a sequence of $M$ natural language utterances
$\bu = (u_1, \ldots, u_M)$,
the task is to generate a program that manipulates the world state
according to the utterances.
Each utterance $u_m$ describes a single action
that transforms the world state $w_{m-1}$ into a new world state $w_m$.
For training,
the system receives weakly supervised examples
with input $x = (\bu, w_0)$
and the target final world state $y = w_M$.

The dataset includes 3 domains:
\textsc{Alchemy}, \textsc{Tangrams}, and \textsc{Scene}.
The description of each domain can be found in \refapp{tokens_table}.
The domains highlight different linguistic phenomena:
\textsc{Alchemy} features ellipsis (e.g., ``throw the rest out'', ``mix'');
\textsc{Tangrams} features anaphora on actions (e.g., ``repeat step 3'', ``bring it back'');
and \textsc{Scene} features anaphora on entities (e.g., ``he moves back'', ``\dots to his left'').
Each domain contains roughly 3,700 training and 900 test examples. Each example
contains 5 utterances and is labeled with the target world state after each
utterance, but not the target program.

\paragraph{Spurious programs.}
Given a training example $(\bu, w_0, w_M)$, our goal is to find the true
underlying program $\bz^*$ which reflects the meaning of $\bu$. The constraint
that $\bz^*$ must transform $w_0$ into $w_M$, i.e. $\bz(w_0) = w_M$, is not
enough to uniquely identify the true $\bz^*$, as there are often many $\bz$
satisfying $\bz(w_0) = w_M$: in our experiments, we found at least 1600 on
average for each example. Almost all do not capture the meaning of $\bu$ (see
\reffig{spurious}). We refer to these incorrect $\bz$'s as \emph{spurious
programs}. Such programs encourage the model to learn an incorrect mapping from
language to program operations: e.g., the spurious program in \reffig{spurious}
would cause the model to learn that ``man in the yellow hat'' maps to
\C{hasShirt(red)}.

\paragraph{Spurious programs in SCONE.}
In this dataset, utterances often reference objects in different ways (e.g.
a person can be referenced by shirt color, hat color, or position). Hence, any target programming language must also
support these different reference strategies. As a result, even a single action
such as moving a person to a target destination can be achieved by many
different programs, each selecting the person and destination in a different
way. Across multiple actions, the number of programs grows
combinatorially.\footnote{The number of well-formed programs in \textsc{Scene}
exceeds $10^{15}$} Only a few programs actually implement the correct
reference strategy as defined by the
utterance. This problem would be more severe in any more general-purpose language
(e.g. Python).

 \section{Model}
\label{subsec:model}

We formulate program generation as a sequence prediction problem.
We represent a program as a sequence of program tokens in postfix notation;
for example,
\C{move(hasHat(yellow), leftOf(hasShirt(blue)))}
is linearized as
\C{yellow~hasHat~blue~hasShirt~leftOf move}.
This representation
also allows us to incrementally execute programs from left to right using a stack:
constants (e.g., \C{yellow}) are pushed onto the stack,
while functions (e.g., \C{hasHat}) pop appropriate arguments from the stack
and push back the computed result (e.g., the list of people with yellow hats).
\refapp{tokens_table} lists the full set of program tokens, $\mathcal{Z}$, and how they are executed.
Note that each action always ends with an \emph{action token} (e.g., \C{move}).

Given an input $x = (\bu, w_0)$,
the model generates program tokens $z_1, z_2, \dots$
from left to right
using a neural encoder-decoder model with attention \cite{bahdanau2015neural}.
Throughout the generation process,
the model maintains an utterance pointer, $m$, initialized to 1.
To generate $z_t$, the model's encoder first encodes the utterance $u_m$
into a vector $e_m$.
Then, based on $e_m$ and previously generated tokens $z_{1:t-1}$,
the model's decoder defines a distribution $p(z_t\mid x, z_{1:t-1})$ over the possible values of $z_t \in \mathcal{Z}$.
The next token $z_t$ is sampled from this distribution.
If an action token (e.g., \C{move}) is generated,
the model increments the utterance pointer $m$.
The process terminates when all $M$ utterances are processed.
The final probability of generating a particular program $\bz = (z_1,\dots,z_T)$
is $p(\bz\mid x) = \prod_{t=1}^T p(z_t\mid x,z_{1:t-1})$.

\paragraph{Encoder.}
The utterance $u_m$ under the pointer is encoded using a bidirectional LSTM:
\begin{align*}
h_i^\mathrm{F} &= \operatorname{LSTM}(h_{i-1}^\mathrm{F}, \Phi_\mathrm{u}(u_{m,i})) \\
h_i^\mathrm{B} &= \operatorname{LSTM}(h_{i+1}^\mathrm{B}, \Phi_\mathrm{u}(u_{m,i})) \\
h_i &= [h_i^\mathrm{F}; h_i^\mathrm{B}],
\end{align*}
where $\Phi_\mathrm{u}(u_{m,i})$ is the fixed GloVe word embedding \cite{pennington2014glove} of the $i$th word in $u_m$.
The final utterance embedding is the concatenation
$e_m = [h_{|u_m|}^\mathrm{F}; h_1^\mathrm{B}]$.

\paragraph{Decoder.}
Unlike \newcite{bahdanau2015neural},
which used a recurrent network for the decoder,
we opt for a feed-forward network for simplicity.
We use $e_m$ and an embedding $f(z_{1:t-1})$ of the previous execution history
(described later)
as inputs to compute an attention vector $c_t$:
\begin{align*}
q_t &= \operatorname{ReLU}(W_\mathrm{q}[e_m; f(z_{1:t-1})]) \\
\alpha_i &\propto \exp(q_t^\top W_\mathrm{a} h_i) \qquad (i=1,\dots,|u_m|)\\
c_t &= \sum_i \alpha_i h_i.
\end{align*}

Finally, after concatenating $q_t$ with $c_t$,
the distribution over the set $\sZ$ of possible program tokens is computed
via a softmax:
\[p(z_{t}\mid x,z_{1:t-1})\propto\exp(\Phi_\mathrm{z}(z_{t})^{\top}W_\mathrm{s}[q_{t};c_{t}]),\]
where $\Phi_\mathrm{z}(z_t)$ is the embedding for token $z_t$.

\paragraph{Execution history embedding.}
We compare two options for $f(z_{1:t-1})$, our embedding of the execution
history. A standard approach is to simply take the $k$ most recent tokens
$z_{t-k:t-1}$ and concatenate their embeddings. We will refer to this as
\textsc{Tokens} and use $k=4$ in our experiments.

We also consider a new approach which leverages our ability to incrementally
execute programs using a stack. We summarize the execution
history by embedding the state of the stack at time $t-1$, achieved by
concatenating
the embeddings of all values on the stack. (We limit the maximum stack size to 3.)
We refer to this as
\textsc{Stack}.

 \section{Reinforcement learning versus maximum marginal likelihood}
\label{sec:rl_vs_lv}

Having formulated our task as a sequence prediction problem, we must still
choose a learning algorithm.
We first compare two standard paradigms: reinforcement learning (RL) and
maximum marginal likelihood (MML). In the next section, we propose a better
alternative.

\subsection{Comparing objective functions}

\paragraph{Reinforcement learning.}
From an RL perspective, given a training example $(x, y)$,
a policy makes a sequence of decisions
$\bz = (z_1, \ldots, z_T)$, and then receives a reward at the end of the
episode: $R(\bz) = 1$ if $\bz$ executes to $y$ and 0 otherwise
(dependence on $x$ and $y$ has been omitted from the notation).

We focus on \emph{policy gradient} methods, in which a stochastic policy
function is trained to maximize the expected reward. In our setup, $p_{\theta}(\bz
\mid x)$ is the policy (with parameters $\theta$), and its expected reward on a given example $(x,
y)$ is
\begin{equation} \label{eqn:rlG}
G(x, y) = \sum_\bz R(\bz)\,p_\theta(\bz\mid x),
\end{equation}
where the sum is over all possible programs.
The overall RL objective, $\JRL$, is the expected reward
across examples:
\begin{equation} \label{eqn:rl}
\JRL = \sum_{(x,y)} G(x, y).
\end{equation}

\paragraph{Maximum marginal likelihood.}
The MML perspective assumes that $y$ is
generated by a partially-observed random process: conditioned on $x$, a latent
program $\bz$ is generated, and conditioned on $\bz$, the observation $y$
is generated. This implies the marginal likelihood:
\begin{equation} \label{eqn:mmlMarginal}
p_\theta(y\mid x) = \sum_\bz p(y\mid \bz)\,p_\theta(\bz\mid x).
\end{equation}
Note that since the execution of $\bz$ is deterministic, $p_\theta(y \mid \bz) =
1$ if $\bz$ executes to $y$ and 0 otherwise.
The log marginal likelihood of the data is then
\begin{align}
\JMML &= \log \LMML, \label{eqn:mml} \\
\text{\hspace{-2em}where\quad} \LMML &= \prod_{(x, y)} p_\theta(y\mid x). \label{eqn:mmlProduct}
\end{align}
To estimate our model parameters $\theta$, we maximize $\JMML$ with respect to
$\theta$.

With our choice of reward, the RL expected reward \refeqn{rlG}
is equal to the MML marginal probability \refeqn{mmlMarginal}.
Hence the only difference between the two formulations is that
in RL we optimize the \emph{sum} of expected rewards \refeqn{rl},
whereas in MML we optimize the \emph{product} \refeqn{mmlProduct}.\footnote{
Note that the log of the product in \refeqn{mmlProduct} does \emph{not}
equal the sum in \refeqn{rl}.}

\subsection{Comparing gradients}

In both policy gradient and MML, the objectives are typically optimized via
(stochastic) gradient ascent.
The gradients of $\JRL$ and $\JMML$ are closely related. They both have the
form:
\begin{align} \label{eqn:grad}
\nabla_{\theta}J &= \sum_{(x,y)} \mathbb E_{\bz \sim q}\left[ R(\bz) \nabla
\log \policy \right] \\
&= \sum_{(x,y)}\sum_{\bz} q(\bz) R(\bz) \nabla \log \policy, \nonumber
\end{align}
where $q(\bz)$ equals
\begin{align}
q_{\mathrm{RL}}(\bz) &= \policy \quad \mbox{for $\JRL$,} \label{eqn:grad_rl} \\
q_{\mathrm{MML}}(\bz) &= \frac{R(\bz) \policy}{\sum_{\tilde{\bz}} R(\tilde{\bz}) p_\theta (\tilde{\bz} \mid x)} \label{eqn:grad_mml} \\
&= p_\theta(\bz \mid x, R(\bz) \neq 0) \quad \mbox{for $\JMML$.} \nonumber
\end{align}

Taking a step in the direction of $\nabla \log \policy$ upweights the
probability of $\bz$, so we can heuristically think of the gradient as
attempting to upweight each reward-earning program $\bz$ by a \emph{gradient
weight} $q(\bz)$. In \refsubsec{meritocratic}, we argue why $q_{\mathrm{MML}}$ is
better at guarding against spurious programs, and propose an even better alternative.

\subsection{Comparing gradient approximation strategies}

It is often intractable to compute the gradient \refeqn{grad}
because it involves taking an expectation over all possible programs.
So in practice, the expectation is approximated.

In the policy gradient literature, \emph{Monte Carlo integration} (MC) is the
typical approximation strategy. For example, the popular REINFORCE algorithm \citep{williams1992simple}
uses Monte Carlo sampling to compute an unbiased estimate of the gradient:
\begin{equation}
\Delta_{\mathrm{MC}} = \frac{1}{B} \sum_{\bz \in \mathcal{S}} [R(\bz) - c] \nabla \log \policy,
\end{equation}
where $\mathcal{S}$ is a collection of $B$ samples $\bz^{(b)} \sim q(\bz)$, and 
$c$ is a \emph{baseline} \citep{williams1992simple} used to reduce the variance of the
estimate without altering its expectation.

In the MML literature for latent sequences, the expectation is typically
approximated via \emph{numerical integration} (NUM) instead:
\begin{equation}
\Delta_{\mathrm{NUM}} = \sum_{\bz \in \mathcal{S}} q(\bz) R(\bz) \nabla \log \policy.
\end{equation}
where the programs in $\mathcal{S}$ come from beam search.

\paragraph{Beam search.}
Beam search generates a set of programs via the following process. At
step $t$ of beam search, we maintain a beam $\mathcal{B}_t$ of at most $B$
\emph{search states}. Each state $s \in \mathcal{B}_t$ represents a partially
constructed program, $s = (z_1, \ldots, z_t)$ (the first $t$ tokens of the
program). For each state $s$ in the beam, we generate all possible
\emph{continuations},
\begin{align*}
\C{cont}(s) &= \C{cont}((z_1, \ldots, z_t)) \\
&= \left\{ (z_1, \ldots, z_t, z_{t+1}) \mid z_{t+1} \in \mathcal{Z} \right\}.
\end{align*}
We then take the union of these continuations, $\C{cont}(\mathcal{B}_t) =
\bigcup_{s \in \mathcal{B}_t} \C{cont}(s)$. The new beam $\mathcal{B}_{t+1}$ is
simply the highest scoring $B$ continuations in $\C{cont}(\mathcal{B}_t)$, as
scored by the policy, $p_\theta(s \mid x)$. Search is halted after a fixed number of iterations or
when there are no continuations possible.
$\mathcal{S}$ is then the set of all complete programs discovered during
beam search. We will refer to this as \emph{beam search MML} (BS-MML).

In both policy gradient and MML, we think of the procedure used to produce
the set of programs $\mathcal{S}$ as an \emph{exploration strategy} which
searches for programs that produce reward.
One advantage of numerical integration is that it allows us to decouple the
\emph{exploration strategy} from the \emph{gradient weights} assigned to
each program.
 \section{Tackling spurious programs}
\label{sec:spurious}

In this section, we illustrate why spurious programs are problematic for the
most commonly used methods in RL (REINFORCE) and MML (beam search MML). We
describe two key problems and propose a solution to each, based on insights
gained from our comparison of RL and MML in \refsec{rl_vs_lv}.

\subsection{Spurious programs bias exploration}

As mentioned in \refsec{rl_vs_lv}, REINFORCE and BS-MML both employ an
\emph{exploration strategy} to approximate their respective gradients. In both
methods, exploration is guided by the current model policy, whereby programs
with high probability under the current policy are more likely to be explored. A
troubling implication is that programs with low probability under the current
policy are likely to be overlooked by exploration.

If the current policy incorrectly assigns low probability to the correct program
$\bz^*$, it will likely fail to discover $\bz^*$ during exploration, and will
consequently fail to upweight the probability of $\bz^*$. This repeats on every gradient step,
keeping the probability of $\bz^*$ perpetually low. The same
feedback loop can also cause already high-probability spurious programs to gain
even more probability. From this, we see that exploration is sensitive to
initial conditions: the rich get richer, and the poor get poorer.

Since there are often thousands of spurious programs and only a few correct
programs, spurious programs are usually found first.
Once spurious programs get a head start, exploration increasingly biases
towards them.

\begin{figure}\center
\includegraphics[width=\columnwidth]{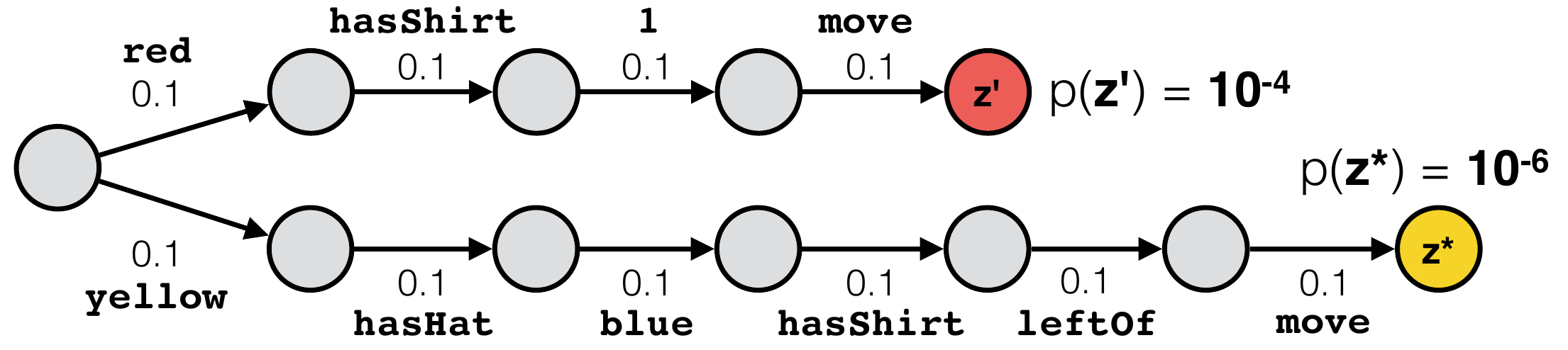}
\caption{
Two possible paths in the tree of all possible programs. One path leads to
the spurious program $\bz^\prime$ (red) while the longer path leads to the correct program $\bz^*$ (gold).
Each edge represents a decision and shows
the probability of that decision under a uniform policy.
The shorter program has two orders of magnitude higher probability.}
\label{fig:search_tree}
\end{figure}

As a remedy, one could try initializing parameters such that
the model puts a uniform distribution over all possible programs. A seemingly
reasonable tactic is to initialize parameters such that the model policy
puts near-uniform probability over the decisions at each time step.
However, this causes shorter programs to have orders of magnitude higher
probability than longer programs, as illustrated in \reffig{search_tree} and as
we empirically observe. A more sophisticated approach might
involve approximating the total number of programs reachable from each point in
the program-generating decision tree. However, we instead propose to
reduce sensitivity to the initial distribution over programs.

\subsubsection*{Solution: randomized beam search}
\label{subsec:randomized}

One solution to biased exploration is to simply rely less on the untrustworthy
current policy. We can do this by injecting random noise into exploration.

In REINFORCE, a common solution is to sample from an $\epsilon$-greedy
variant of the current policy.
On the other hand,
MML exploration with beam search is deterministic. However, it has
a key advantage over REINFORCE-style sampling: even if one program occupies almost all
probability under the current policy (a peaky distribution), beam search will still use
its remaining beam capacity to explore at least $B-1$ other programs. In contrast, sampling
methods will repeatedly visit the mode of the distribution.

To get the best of both worlds, we propose a simple \emph{$\epsilon$-greedy randomized beam search}.
Like regular beam search, at iteration $t$ we compute the set of all
continuations $\C{cont}(\mathcal{B}_t)$ and sort them by their model probability $p_\theta(s \mid x)$.
But instead of selecting the $B$ highest-scoring continuations,
we choose $B$ continuations one by one without replacement from $\C{cont}(\mathcal{B}_t)$.
When choosing a continuation from the remaining pool,
we either uniformly sample a random continuation
with probability $\epsilon$,
or pick the highest-scoring continuation in the pool
with probability $1 - \epsilon$.
Empirically, we find that this performs much better than both
classic beam search and $\epsilon$-greedy sampling (\reftab{x-exploration}).

\subsection{Spurious programs dominate gradients}

In both RL and MML, even if exploration is perfect and the gradient is exactly
computed, spurious programs can still be problematic.

Even if perfect exploration visits every program, we see from the gradient weights $q(\bz)$
in \refeqn{grad_rl} and \refeqn{grad_mml} that
programs are weighted proportional to their current policy probability.
If a spurious program $\bz^\prime$ has 100
times higher probability than $\bz^*$ as in \reffig{search_tree}, the gradient
will spend roughly 99\% of its magnitude upweighting towards $\bz^\prime$ and
only 1\% towards $\bz^*$ even though the two programs get the same reward.

This implies that it would take many updates for $\bz^*$ to catch up. In fact,
$\bz^*$ may never catch up, depending on the gradient updates for other
training examples.
 Simply
increasing the learning rate is inadequate, as it would cause the
model to take overly large steps towards $\bz^\prime$, potentially causing
optimization to diverge.

\subsubsection*{Solution: the meritocratic update rule}
\label{subsec:meritocratic}

To solve this problem, we want the upweighting to be more
``meritocratic'': any program that obtains reward should be upweighted roughly equally.

We first observe that $\JMML$ already improves over $\JRL$ in this regard. From \refeqn{grad},
we see that the gradient weight $q_{\mathrm{MML}}(\bz)$ is the policy distribution
restricted to and renormalized over only reward-earning programs. This
renormalization makes the gradient weight uniform \emph{across examples}:
even if all reward-earning programs for a particular example have
very low model probability, their combined gradient weight
$\sum_\bz q_{\mathrm{MML}}(\bz)$ is always 1. In our experiments,
$\JMML$ performs significantly better than $\JRL$ (\reftab{x-beta}).

However, while $\JMML$ assigns uniform weight across examples, it is still not
uniform
over the programs \emph{within each example}. Hence we propose a new update rule
which goes one step further in pursuing uniform updates. Extending
$q_{\mathrm{MML}}(\bz)$, we define a $\beta$-smoothed version:
\begin{equation}
q_\beta(\bz)=\frac{q_{\mathrm{MML}}(\bz)^{\beta}}{\sum_{\tilde{\bz}}q_{\mathrm
{MML}}(\tilde{\bz})^{\beta}}.
\end{equation}

When $\beta = 0$, our weighting is completely uniform across all reward-earning programs within
an example
while $\beta = 1$ recovers the original MML weighting. Our new
update rule is to simply take a modified gradient step where $q =
q_\beta$.\footnote{
Also, note that if exploration were exhaustive, $\beta=0$ would be equivalent to supervised learning
using the set of all reward-earning programs as targets.
}
We will refer to this as the $\beta$-\emph{meritocratic} update rule.

\subsection{Summary of the proposed approach}

\begin{table*}[th]
\center\small
\begin{tabular}{cccc}
\textbf{Method} & \textbf{Approximation of $E_{q}\left[\cdot\right]$} & \textbf{Exploration strategy} & \textbf{Gradient weight $q(\bz)$}\tabularnewline
\hline 
REINFORCE & Monte Carlo integration & independent sampling & $\policy$\tabularnewline
BS-MML & numerical integration & beam search & $p_\theta(\bz \mid x,R(\bz)\neq0)$\tabularnewline
\ourmodel & numerical integration & randomized beam search & $q_\beta(\bz)$\tabularnewline
\hline 
\end{tabular}
\caption{
\ourmodel combines qualities of both REINFORCE (RL) and BS-MML.
For approximating the expectation over $q$ in the gradient, we use numerical integration as in BS-MML.
Our exploration strategy is a hybrid of search (MML) and off-policy sampling (RL).
Our gradient weighting is equivalent to MML when $\beta=1$ and more ``meritocratic'' than both MML and REINFORCE for
lower values of $\beta$.
}
\label{tab:x-design-space}
\end{table*}

We described two problems\footnote{
These problems concern the gradient w.r.t. a single
example. The full gradient averages over multiple examples, which helps
separate correct from spurious.
E.g., if multiple examples all mention ``yellow hat'',
we will find a correct program parsing this as \C{hasHat(yellow)} for each
example, whereas the spurious programs we find will follow no consistent pattern.
Consequently, spurious gradient contributions may cancel out while
correct program gradients will all ``vote'' in the same direction.
}
and their solutions: we reduce exploration bias using \emph{$\epsilon$-greedy
randomized beam search} and perform more balanced optimization using the
\emph{$\beta$-meritocratic parameter update rule}. We call our resulting
approach \ourmodel. \reftab{x-design-space} summarizes how \ourmodel combines
desirable qualities from both REINFORCE and BS-MML.

 \section{Experiments}

\paragraph{Evaluation.}
We evaluate our proposed methods on all three domains of the SCONE dataset.
Accuracy is defined as the
percentage of test examples where the model produces the correct final world
state $w_M$. All test examples have $M=5$ (5utts), but we also report accuracy
after processing the first 3 utterances (3utts).
To control for the effects of randomness, we train 5 instances of each model
with different random seeds. We report the median accuracy of the
instances unless otherwise noted.

\paragraph{Training.}
Following \citet{long2016projections}, we decompose each training
example into smaller examples. Given an example with 5 utterances, $\bu = [u_1,
\ldots, u_5]$, we consider all length-1 and length-2 substrings of $\bu$:
$[u_1], [u_2], \ldots, [u_3, u_4], [u_4, u_5]$ (9 total). We form a new training
example from each substring, e.g., $(\bu^\prime, w_0^\prime, w_M^\prime)$ where
$\bu^\prime = [u_4, u_5]$, $w_0^\prime=w_3$ and $w_M^\prime=w_5$.

All models are implemented in TensorFlow \citep{abadi2015tensorflow}.
Model parameters are randomly initialized \citep{glorot2010understanding}, with no pre-training.
We use the Adam
optimizer \citep{kingma2014adam} (which is applied to the gradient in
\refeqn{grad}), a learning rate of 0.001, a mini-batch size of 8
examples (different from the beam size), and train until accuracy on the
validation set converges (on average about 13,000 steps). We use fixed GloVe
vectors \citep{pennington2014glove} to embed the words in each utterance.

\paragraph{Hyperparameters.}
For all models, we performed a grid search over hyperparameters to maximize
accuracy on the validation set. Hyperparameters include the learning rate, the
baseline in REINFORCE, $\epsilon$-greediness and $\beta$-meritocraticness.
For REINFORCE, we also experimented with a regression-estimated baseline \citep{ranzato2015sequence},
but found it to perform worse than a constant baseline.

\subsection{Main results}

\paragraph{Comparison to prior work.}
\reftab{x-final} compares \ourmodel to results from
\citet{long2016projections} as well as two baselines, REINFORCE and BS-MML
(using the same neural model but different learning algorithms). Our
approach achieves new state-of-the-art results by a significant margin,
especially on the \textsc{Scene} domain, which features the most complex program
syntax. We report the results for REINFORCE, BS-MML, and \ourmodel on the seed
and hyperparameters that achieve the best validation accuracy.

We note that REINFORCE performs very well on \textsc{Tangrams} but
worse on \textsc{Alchemy} and very poorly on \textsc{Scene}. This might be
because the program syntax for \textsc{Tangrams} is simpler than the other two: there
is no other way to refer to objects except by index.

We also found that REINFORCE required $\epsilon$-greedy exploration to make any
progress. Using $\epsilon$-greedy greatly skews the Monte Carlo approximation of
$\nabla \JRL$, making it more uniformly weighted over programs in a similar
spirit to using $\beta$-meritocratic gradient weights $q_\beta$. However,
$q_\beta$ increases uniformity over reward-earning programs only, rather than
over all programs.

\begin{table}[t]\centering\small
  \begin{tabular}{lc@{\;\;}cc@{\;\;}cc@{\;\;}c}
  & \multicolumn{2}{c}{\textsc{Alchemy}}
  & \multicolumn{2}{@{}c@{}}{\textsc{Tangrams}}
  & \multicolumn{2}{c}{\textsc{Scene}} \\
    system & 3utts & 5utts & 3utts & 5utts & 3utts & 5utts \\ \hline
  \textsc{Long+16} & 56.8 & 52.3 & 64.9 & 27.6 & 23.2 & 14.7 \\
  REINFORCE & 58.3 & 44.6 & \textbf{68.5} & \textbf{37.3} & 47.8 & 33.9 \\
  BS-MML & 58.7 & 47.3 & 62.6 & 32.2 & 53.5 & 32.5 \\
  \ourmodel & \bf{66.9} & \bf{52.9} & 65.8 & 37.1 &
                       \bf{64.8} & \bf{46.2} \\
  \hline
\end{tabular}
\caption{\textbf{Comparison to prior work.}
\textsc{Long+16} results are directly from \citet{long2016projections}. Hyperparameters are chosen by best performance on validation set (see \refapp{hyperparams}).
\vspace*{-.3em}
}\label{tab:x-final}
\end{table}

\paragraph{Effect of randomized beam search.}
\reftab{x-exploration} shows that $\epsilon$-greedy randomized beam search
consistently outperforms classic beam search.
Even when we increase the beam size of classic beam search to 128, it still
does not surpass randomized beam search with a beam of 32, and further increases
yield no additional improvement.

\begin{table}[t]\centering\small
  \begin{tabular}{@{}l@{}cc@{\;\;}cc@{\;\;}cc@{\;\;}c}
  & & \multicolumn{2}{c}{\textsc{Alchemy}}
  & \multicolumn{2}{@{}c@{}}{\textsc{Tangrams}}
  & \multicolumn{2}{c}{\textsc{Scene}} \\
  random & beam & 3utts & 5utts & 3utts & 5utts & 3utts & 5utts \\ \hline
  \multicolumn{8}{c}{\textbf{classic beam search}} \\
   None & 32 & 30.3 & 23.2 & 0.0 & 0.0 & 33.4 & 20.1 \\
   None & 128 & 59.0 & 46.4 & 60.9 & 28.6 & 24.5 & 13.9 \\
   \hline
  \multicolumn{8}{c}{\textbf{randomized beam search}} \\
         
   $\epsilon = 0.05$ & 32 & 58.7 & 45.5 & 61.1 & 32.5 & 33.4 & 23.0 \\
   $\epsilon = 0.15$ & 32 & \bf{61.3} & 48.3 & \bf{65.2} & \bf{34.3} & 50.8 & 33.5 \\
   $\epsilon = 0.25$ & 32 & 60.5 & \bf{48.6} & 60.0 & 27.3 & \bf{54.1} &
   \bf{35.7} \\
  \hline
\end{tabular}
\caption{\textbf{Randomized beam search.}
All listed models use gradient weight $q_{\mathrm{MML}}$ and \textsc{Tokens} to represent execution history.
}
\label{tab:x-exploration}
\end{table}

\paragraph{Effect of $\beta$-meritocratic updates.}
\reftab{x-beta} evaluates the impact of $\beta$-meritocratic parameter updates
(gradient weight $q_\beta$). More uniform upweighting across reward-earning
programs leads to higher accuracy and fewer spurious programs, especially in
\textsc{Scene}. However, no single value of $\beta$ performs best over all
domains.

\begin{figure}\center
\includegraphics[width=\columnwidth]{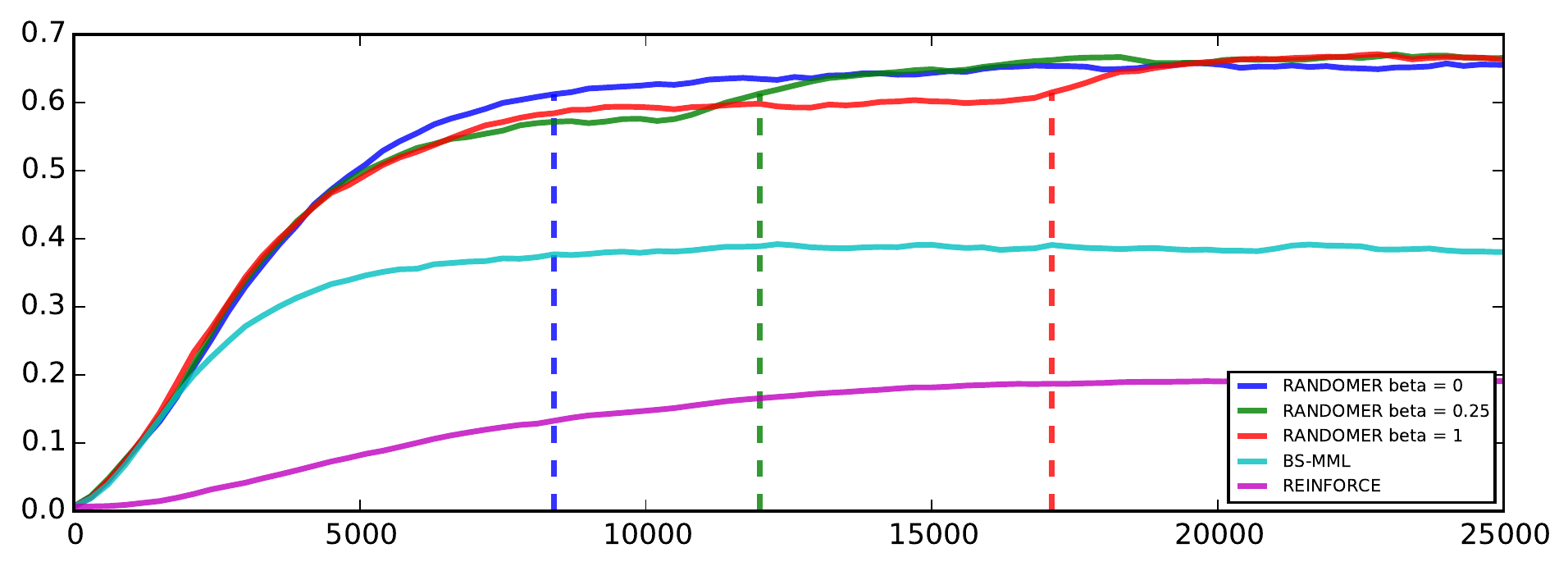}
\caption{
Validation set accuracy ($y$-axis) across training iterations ($x$-axis) on
\textsc{Alchemy}. We compare \ourmodel, BS-MML and REINFORCE. Vertical lines
mark the first time each model surpasses 60\% accuracy. \ourmodel with $\beta=0$
reaches this point twice as fast as $\beta=1$. REINFORCE plateaus for a long time,
then begins to climb after 40k iterations (not shown). Training runs are averaged over 5 seeds.}
\label{fig:faster_training}
\end{figure}

Choosing the right value of $\beta$ in \ourmodel significantly accelerates
training. \reffig{faster_training} illustrates that
while $\beta=0$ and $\beta=1$ ultimately achieve similar accuracy on
\textsc{Alchemy}, $\beta=0$ reaches good performance in half the time.

Since lowering $\beta$ reduces trust in the model policy,
$\beta < 1$ helps in early training when the current policy is untrustworthy.
However, as it grows more trustworthy, $\beta < 1$
begins to pay a price for ignoring it. Hence, it may be worthwhile to anneal
$\beta$ towards 1 over time.

\begin{table}[t]\centering\small
\begin{tabular}{cc@{\;\;}cc@{\;\;}cc@{\;\;}c}
  & \multicolumn{2}{c}{\textsc{Alchemy}}
  & \multicolumn{2}{@{}c@{}}{\textsc{Tangrams}}
  & \multicolumn{2}{c}{\textsc{Scene}} \\
  $q(\bz)$ & 3utts & 5utts & 3utts & 5utts & 3utts & 5utts \\ \hline
  $q_{\mathrm{RL}}$ & 0.2 & 0.0 & 0.9 & 0.6 & 0.0 & 0.0 \\
  $q_{\mathrm{MML}}\,(q_{\beta=1})$ & 61.3 & 48.3 & \bf{65.2} & \bf{34.3} &
  50.8 & 33.5 \\ \hline
  $q_{\beta=0.25}$ & \bf{64.4} & \bf{48.9} & 60.6 & 29.0 & 42.4 &
  29.7 \\
  $q_{\beta=0}$ & 63.6 & 46.3 & 54.0 & 23.5 & \bf{61.0} & \bf{42.4} \\
  \hline
\end{tabular}
\caption{
\textbf{\boldmath $\beta$-meritocratic updates.} All listed models use
randomized beam search, $\epsilon=0.15$ and \textsc{Tokens} to represent execution history.
}
\label{tab:x-beta}
\end{table}

\paragraph{Effect of execution history embedding.} 
\reftab{x-stack} compares our two proposals for embedding the execution history:
\textsc{Tokens} and \textsc{Stack}. \textsc{Stack} performs better in the two
domains where an object can be referenced in multiple ways (\textsc{Scene} and \textsc{Alchemy}).
\textsc{Stack} directly embeds objects on the stack, invariant to the way
in which they were pushed onto the stack, unlike \textsc{Tokens}. We hypothesize that
this invariance increases robustness to spurious behavior: if a program
accidentally pushes the right object onto the stack via spurious means, the
model can still learn the remaining steps of the program without conditioning on
a spurious history.

\begin{table}[t]\centering\small
  \begin{tabular}{lc@{\;\;}cc@{\;\;}cc@{\;\;}c}
  & \multicolumn{2}{c}{\textsc{Alchemy}}
  & \multicolumn{2}{@{}c@{}}{\textsc{Tangrams}}
  & \multicolumn{2}{c}{\textsc{Scene}} \\
  & 3utts & 5utts & 3utts & 5utts & 3utts & 5utts \\ \hline
  \textsc{History} & 61.3 & 48.3 & \bf{65.2} & \bf{34.3} & 50.8 & 33.5 \\
  \textsc{Stack} & \bf{64.2} & \bf{53.2} & 63.0 & 32.4 & \bf{59.5} & \bf{43.1} \\
  \hline
\end{tabular}
\caption{\textbf{\textsc{Tokens} vs \textsc{Stack} embedding.}
Both models use
$\epsilon=0.15$ and gradient weight $q_{\mathrm{MML}}$.
}
\label{tab:x-stack}
\end{table}

\paragraph{Fitting vs overfitting the training data.}
\reftab{x-beam} reveals that BS-MML and \ourmodel use different strategies to
fit the training data. On the depicted training example, BS-MML
actually achieves higher expected reward / marginal probability than \ourmodel, but it
does so by putting most of its probability on a spurious program---a form of
overfitting. In contrast, \ourmodel spreads probability mass over multiple
reward-earning programs, including the correct ones.

As a consequence of overfitting, we observed at test time that BS-MML only
references people by positional indices instead of by shirt or hat color,
whereas \ourmodel successfully learns to use multiple reference strategies.

\newcommand{\CC}[2]{\parbox[t]{8cm}{\C{#1\\ \hspace*{3em}#2}}}
\newcommand{\bCorrect}{\C{*}}
\newcommand{\bSpurious}{\C{o}}
\newcommand{\bWrong}{\C{x}}

\begin{table}[t]\small
\textbf{Utterance:} the man in the purple shirt and red hat moves just to the right of the man in the red shirt and yellow hat
\\[1ex]
{\centering
  \begin{tabular}{@{}c@{\hspace{.2cm}}p{6.5cm}@{}r@{}}
  & program & \hspace{-.2in}prob \\ \hline
  \multicolumn{3}{c}{\textbf{\ourmodel} ($\epsilon = 0.15$, $\beta = 0$)} \\
    \bCorrect & \CC{move(hasHat(red),}{rightOf(hasHat(red)))} &  0.122 \\
    \bCorrect & \CC{move(hasShirt(purple),}{rightOf(hasShirt(red)))} & 0.061  \\
   \bSpurious & \CC{move(hasHat(red),}{rightOf(index(allPeople, 1)))} &  0.059 \\
    \bCorrect & \CC{move(hasHat(red),}{rightOf(hasHat(yellow)))} &  0.019 \\
   \bSpurious & \CC{move(index(allPeople, 2),}{rightOf(hasShirt(red)))} &  0.018 \\
      \bWrong & \C {move(hasHat(red), 8)} &  0.018 \\
  \hline \multicolumn{3}{c}{\textbf{BS-MML}} \\
   \bSpurious & \C{move(index(allPeople, 2), 2)} &  0.887 \\
      \bWrong & \C{move(index(allPeople, 2), 6)} &  0.041 \\
      \bWrong & \C{move(index(allPeople, 2), 5)} &  0.020 \\
      \bWrong & \C{move(index(allPeople, 2), 8)} &  0.016 \\
      \bWrong & \C{move(index(allPeople, 2), 7)} &  0.009 \\
      \bWrong & \C{move(index(allPeople, 2), 3)} &  0.008 \\
  \hline
  \end{tabular}
}
\caption{Top-scoring predictions for a \emph{training} example
from \textsc{Scene}
(* = correct, o = spurious, x = incorrect).
\ourmodel distributes probability mass
over numerous reward-earning programs (including the correct ones), while
classic beam search MML overfits to one spurious program, giving it very high
probability.
}\label{tab:x-beam}
\end{table}

 \section{Related work and discussion}

\textbf{Semantic parsing from indirect supervision.}
Our work is motivated by the classic problem of learning semantic parsers from
indirect supervision
\citep{clarke10world,liang11dcs,artzi11conversations,artzi2013weakly,reddy2014large,pasupat2015compositional}.
We are interested in the initial stages of training from scratch,
where getting any training signal is difficult due to the combinatorially large search space.
We also highlighted the problem of spurious programs which capture reward but give incorrect generalizations.

Maximum marginal likelihood with beam search (BS-MML) is traditionally used to learn
semantic parsers from indirect supervision.

\textbf{Reinforcement learning.} 
Concurrently, there has been a recent surge of interest in reinforcement
learning, along with the wide application of the classic REINFORCE algorithm
\citep{williams1992simple}---to troubleshooting
\citep{branavan09reinforcement}, dialog generation \citep{li2016rl},
game playing
\citep{narasimhan2015language}, coreference resolution
\citep{clark2016deep}, machine translation \citep{norouzi2016reward}, and even
semantic parsing \citep{liang2017nsm}. Indeed, the challenge of training
semantic parsers from indirect supervision is perhaps better captured by the
notion of sparse rewards in reinforcement learning.

The RL answer would be better exploration, which can take many forms including
simple action-dithering such as $\epsilon$-greedy, entropy regularization
\citep{williams1991function}, Monte Carlo tree search
\citep{coulom2006efficient}, randomized value functions
\citep{osband2014generalization,osband2016deep}, and methods which prioritize
learning environment dynamics \citep{duff2002optimal} or under-explored states
\citep{kearns2002near,bellemare2016unifying,nachum2016improving}. The majority
of these methods employ Monte Carlo sampling for exploration. In contrast, we find
randomized beam search to be more suitable in our setting, because it explores
low-probability states even when the policy distribution is peaky. Our $\beta$-meritocratic
update also depends on the fact that beam search
returns an \emph{entire set} of reward-earning programs rather than one, since it
renormalizes over the reward-earning set.
While similar to
entropy regularization,
$\beta$-meritocratic update
is more targeted
as it only
increases uniformity of the gradient among reward-earning programs, rather than
across all programs.

Our strategy of using randomized beam search and meritocratic updates lies
closer to MML than RL, but this does not imply that RL has nothing
to offer in our setting. With the simple connection between RL and
MML we established, much of the literature on exploration and variance reduction
in RL can be directly applied to MML problems. Of special interest are methods which incorporate a value function such as actor-critic.

\textbf{Maximum likelihood and RL.} 
It is tempting to group our approach with sequence learning methods which
interpolate between supervised learning and reinforcement learning
\citep{ranzato2015sequence,venkatraman2015improving,ross2011reduction,norouzi2016reward,bengio2015scheduled,levine2014motor}.
These methods generally seek to make RL training easier by pre-training or
``warm-starting'' with fully supervised learning.
This requires each training example to be labeled with a
reasonably correct output sequence. In our setting, this would amount to
labeling each example with the correct program, which is not known. Hence, these
methods cannot be directly applied.

Without access to correct output sequences, we cannot directly maximize
likelihood, and instead resort to maximizing the \emph{marginal} likelihood (MML).
Rather than proposing MML as a form of pre-training, we argue that MML is a
superior substitute for the standard RL objective, and that the
$\beta$-meritocratic update is even better.

\textbf{Simulated annealing.}
Our $\beta$-meritocratic update employs exponential smoothing, which
bears
resemblance to the simulated annealing strategy of \citet
{och2003minimum,smith2006minimum,shen2015minimum}.
However, a key difference is that these methods smooth the objective function
whereas we smooth an expectation in the gradient. To underscore the difference,
we note that fixing $\beta = 0$ in our method (total smoothing) is quite
effective, whereas total smoothing in the simulated annealing methods would
correspond to a completely flat objective function, and an uninformative
gradient of zero everywhere.

\textbf{Neural semantic parsing.}
There has been recent interest in using recurrent
neural networks for semantic parsing, both for modeling logical forms
\citep{dong2016logical,jia2016recombination,liang2017nsm}
and for end-to-end execution \citep{yin2015enquirer,neelakantan2016neural}. We
develop a neural model for the context-dependent setting, which is made
possible by a new stack-based language similar to \citet{riedel2016programming}.

\paragraph{Acknowledgments.}

This work was supported by the NSF Graduate Research Fellowship under No. DGE-114747
and the NSF CAREER Award under No. IIS-1552635.

\paragraph{Reproducibility.}

Our code is made available at
{\scriptsize\url{https://github.com/kelvinguu/lang2program}}.
Reproducible experiments are available at
{\scriptsize\url{https://worksheets.codalab.org/worksheets/0x88c914ee1d4b4a4587a07f36f090f3e5/}}.

\bibliographystyle{acl_natbib}
\bibliography{refdb/all}

\begin{thebibliography}{}
\expandafter\ifx\csname natexlab\endcsname\relax\def\natexlab#1{#1}\fi

\bibitem[{Abadi et~al.(2015)Abadi, Agarwal, Barham, Brevdo, Chen, Citro,
  Corrado, Davis, Dean, Devin, Ghemawat, Goodfellow, Harp, Irving, Isard, Jia,
  Józefowicz, Kaiser, Kudlur, Levenberg, Mané, Monga, Moore, Murray, Olah,
  Schuster, Shlens, Steiner, Sutskever, Talwar, Tucker, Vanhoucke, Vasudevan,
  Viégas, Vinyals, Warden, Wattenberg, Wicke, Yu, and
  Zheng}]{abadi2015tensorflow}
M.~Abadi, A.~Agarwal, P.~Barham, E.~Brevdo, Z.~Chen, C.~Citro, G.~S. Corrado,
  A.~Davis, J.~Dean, M.~Devin, S.~Ghemawat, I.~J. Goodfellow, A.~Harp,
  G.~Irving, M.~Isard, Y.~Jia, R.~Józefowicz, L.~Kaiser, M.~Kudlur,
  J.~Levenberg, D.~Mané, R.~Monga, S.~Moore, D.~G. Murray, C.~Olah,
  M.~Schuster, J.~Shlens, B.~Steiner, I.~Sutskever, K.~Talwar, P.~A. Tucker,
  V.~Vanhoucke, V.~Vasudevan, F.~B. Viégas, O.~Vinyals, P.~Warden,
  M.~Wattenberg, M.~Wicke, Y.~Yu, and X.~Zheng. 2015.
\newblock Tensorflow: Large-scale machine learning on heterogeneous distributed
  systems.
\newblock {\em arXiv preprint arXiv:1603.04467\/} .

\bibitem[{Artzi and Zettlemoyer(2011)}]{artzi11conversations}
Y.~Artzi and L.~Zettlemoyer. 2011.
\newblock Bootstrapping semantic parsers from conversations.
\newblock In {\em Empirical Methods in Natural Language Processing (EMNLP)\/}.
  pages 421--432.

\bibitem[{Artzi and Zettlemoyer(2013)}]{artzi2013weakly}
Y.~Artzi and L.~Zettlemoyer. 2013.
\newblock Weakly supervised learning of semantic parsers for mapping
  instructions to actions.
\newblock {\em Transactions of the Association for Computational Linguistics
  (TACL)\/} 1:49--62.

\bibitem[{Bahdanau et~al.(2015)Bahdanau, Cho, and Bengio}]{bahdanau2015neural}
D.~Bahdanau, K.~Cho, and Y.~Bengio. 2015.
\newblock Neural machine translation by jointly learning to align and
  translate.
\newblock In {\em International Conference on Learning Representations
  (ICLR)\/}.

\bibitem[{Bellemare et~al.(2016)Bellemare, Srinivasan, Ostrovski, Schaul,
  Saxton, and Munos}]{bellemare2016unifying}
M.~Bellemare, S.~Srinivasan, G.~Ostrovski, T.~Schaul, D.~Saxton, and R.~Munos.
  2016.
\newblock Unifying count-based exploration and intrinsic motivation.
\newblock In {\em Advances in Neural Information Processing Systems (NIPS)\/}.
  pages 1471--1479.

\bibitem[{Bengio et~al.(2015)Bengio, Vinyals, Jaitly, and
  Shazeer}]{bengio2015scheduled}
S.~Bengio, O.~Vinyals, N.~Jaitly, and N.~Shazeer. 2015.
\newblock Scheduled sampling for sequence prediction with recurrent neural
  networks.
\newblock In {\em Advances in Neural Information Processing Systems (NIPS)\/}.
  pages 1171--1179.

\bibitem[{Branavan et~al.(2009)Branavan, Chen, Zettlemoyer, and
  Barzilay}]{branavan09reinforcement}
S.~Branavan, H.~Chen, L.~S. Zettlemoyer, and R.~Barzilay. 2009.
\newblock Reinforcement learning for mapping instructions to actions.
\newblock In {\em Association for Computational Linguistics and International
  Joint Conference on Natural Language Processing (ACL-IJCNLP)\/}. pages
  82--90.

\bibitem[{Clark and Manning(2016)}]{clark2016deep}
K.~Clark and C.~D. Manning. 2016.
\newblock Deep reinforcement learning for mention-ranking coreference models.
\newblock {\em arXiv preprint arXiv:1609.08667\/} .

\bibitem[{Clarke et~al.(2010)Clarke, Goldwasser, Chang, and
  Roth}]{clarke10world}
J.~Clarke, D.~Goldwasser, M.~Chang, and D.~Roth. 2010.
\newblock Driving semantic parsing from the world's response.
\newblock In {\em Computational Natural Language Learning (CoNLL)\/}. pages
  18--27.

\bibitem[{Coulom(2006)}]{coulom2006efficient}
R.~Coulom. 2006.
\newblock Efficient selectivity and backup operators in {M}onte-{C}arlo tree
  search.
\newblock In {\em International Conference on Computers and Games\/}. pages
  72--83.

\bibitem[{Dempster et~al.(1977)Dempster, M., and B.}]{demp1977em}
A.~P. Dempster, L.~N. M., and R.~D. B. 1977.
\newblock Maximum likelihood from incomplete data via the {EM} algorithm.
\newblock {\em Journal of the Royal Statistical Society: Series B\/}
  39(1):1--38.

\bibitem[{Dong and Lapata(2016)}]{dong2016logical}
L.~Dong and M.~Lapata. 2016.
\newblock Language to logical form with neural attention.
\newblock In {\em Association for Computational Linguistics (ACL)\/}.

\bibitem[{Duff(2002)}]{duff2002optimal}
M.~O. Duff. 2002.
\newblock {\em Optimal Learning: Computational procedures for Bayes-adaptive
  Markov decision processes\/}.
\newblock Ph.D. thesis, University of Massachusetts Amherst.

\bibitem[{Glorot and Bengio(2010)}]{glorot2010understanding}
X.~Glorot and Y.~Bengio. 2010.
\newblock Understanding the difficulty of training deep feedforward neural
  networks.
\newblock In {\em International Conference on Artificial Intelligence and
  Statistics\/}.

\bibitem[{Jia and Liang(2016)}]{jia2016recombination}
R.~Jia and P.~Liang. 2016.
\newblock Data recombination for neural semantic parsing.
\newblock In {\em Association for Computational Linguistics (ACL)\/}.

\bibitem[{Kearns and Singh(2002)}]{kearns2002near}
M.~Kearns and S.~Singh. 2002.
\newblock Near-optimal reinforcement learning in polynomial time.
\newblock {\em Machine Learning\/} 49(2):209--232.

\bibitem[{Kingma and Ba(2014)}]{kingma2014adam}
D.~Kingma and J.~Ba. 2014.
\newblock Adam: A method for stochastic optimization.
\newblock {\em arXiv preprint arXiv:1412.6980\/} .

\bibitem[{Krishnamurthy and Mitchell(2012)}]{krishnamurthy2012weakly}
J.~Krishnamurthy and T.~Mitchell. 2012.
\newblock Weakly supervised training of semantic parsers.
\newblock In {\em Empirical Methods in Natural Language Processing and
  Computational Natural Language Learning (EMNLP/CoNLL)\/}. pages 754--765.

\bibitem[{Levine(2014)}]{levine2014motor}
S.~Levine. 2014.
\newblock {\em Motor Skill Learning with Local Trajectory Methods\/}.
\newblock Ph.D. thesis, Stanford University.

\bibitem[{Li et~al.(2016)Li, Monroe, Ritter, Jurafsky, Galley, and
  Gao}]{li2016rl}
J.~Li, W.~Monroe, A.~Ritter, D.~Jurafsky, M.~Galley, and J.~Gao. 2016.
\newblock Deep reinforcement learning for dialogue generation.
\newblock In {\em Empirical Methods in Natural Language Processing (EMNLP)\/}.

\bibitem[{Liang et~al.(2017)Liang, Berant, Le, and Lao}]{liang2017nsm}
C.~Liang, J.~Berant, Q.~Le, and K.~D. F.~N. Lao. 2017.
\newblock Neural symbolic machines: Learning semantic parsers on {F}reebase
  with weak supervision.
\newblock In {\em Association for Computational Linguistics (ACL)\/}.

\bibitem[{Liang et~al.(2011)Liang, Jordan, and Klein}]{liang11dcs}
P.~Liang, M.~I. Jordan, and D.~Klein. 2011.
\newblock Learning dependency-based compositional semantics.
\newblock In {\em Association for Computational Linguistics (ACL)\/}. pages
  590--599.

\bibitem[{Long et~al.(2016)Long, Pasupat, and Liang}]{long2016projections}
R.~Long, P.~Pasupat, and P.~Liang. 2016.
\newblock Simpler context-dependent logical forms via model projections.
\newblock In {\em Association for Computational Linguistics (ACL)\/}.

\bibitem[{Nachum et~al.(2016)Nachum, Norouzi, and
  Schuurmans}]{nachum2016improving}
O.~Nachum, M.~Norouzi, and D.~Schuurmans. 2016.
\newblock Improving policy gradient by exploring under-appreciated rewards.
\newblock {\em arXiv preprint arXiv:1611.09321\/} .

\bibitem[{Narasimhan et~al.(2015)Narasimhan, Kulkarni, and
  Barzilay}]{narasimhan2015language}
K.~Narasimhan, T.~Kulkarni, and R.~Barzilay. 2015.
\newblock Language understanding for text-based games using deep reinforcement
  learning.
\newblock {\em arXiv preprint arXiv:1506.08941\/} .

\bibitem[{Neelakantan et~al.(2016)Neelakantan, Le, and
  Sutskever}]{neelakantan2016neural}
A.~Neelakantan, Q.~V. Le, and I.~Sutskever. 2016.
\newblock Neural programmer: Inducing latent programs with gradient descent.
\newblock In {\em International Conference on Learning Representations
  (ICLR)\/}.

\bibitem[{Norouzi et~al.(2016)Norouzi, Bengio, Jaitly, Schuster, Wu, Schuurmans
  et~al.}]{norouzi2016reward}
M.~Norouzi, S.~Bengio, N.~Jaitly, M.~Schuster, Y.~Wu, D.~Schuurmans, et~al.
  2016.
\newblock Reward augmented maximum likelihood for neural structured prediction.
\newblock In {\em Advances In Neural Information Processing Systems\/}. pages
  1723--1731.

\bibitem[{Och(2003)}]{och2003minimum}
F.~J. Och. 2003.
\newblock Minimum error rate training in statistical machine translation.
\newblock In {\em Association for Computational Linguistics (ACL)\/}. pages
  160--167.

\bibitem[{Osband et~al.(2016)Osband, Blundell, Pritzel, and
  Roy}]{osband2016deep}
I.~Osband, C.~Blundell, A.~Pritzel, and B.~V. Roy. 2016.
\newblock Deep exploration via bootstrapped {DQN}.
\newblock In {\em Advances In Neural Information Processing Systems\/}. pages
  4026--4034.

\bibitem[{Osband et~al.(2014)Osband, Roy, and Wen}]{osband2014generalization}
I.~Osband, B.~V. Roy, and Z.~Wen. 2014.
\newblock Generalization and exploration via randomized value functions.
\newblock {\em arXiv preprint arXiv:1402.0635\/} .

\bibitem[{Pasupat and Liang(2015)}]{pasupat2015compositional}
P.~Pasupat and P.~Liang. 2015.
\newblock Compositional semantic parsing on semi-structured tables.
\newblock In {\em Association for Computational Linguistics (ACL)\/}.

\bibitem[{Pasupat and Liang(2016)}]{pasupat2016inferring}
P.~Pasupat and P.~Liang. 2016.
\newblock Inferring logical forms from denotations.
\newblock In {\em Association for Computational Linguistics (ACL)\/}.

\bibitem[{Pennington et~al.(2014)Pennington, Socher, and
  Manning}]{pennington2014glove}
J.~Pennington, R.~Socher, and C.~D. Manning. 2014.
\newblock Glove: Global vectors for word representation.
\newblock In {\em Empirical Methods in Natural Language Processing (EMNLP)\/}.

\bibitem[{Ranzato et~al.(2015)Ranzato, Chopra, Auli, and
  Zaremba}]{ranzato2015sequence}
M.~Ranzato, S.~Chopra, M.~Auli, and W.~Zaremba. 2015.
\newblock Sequence level training with recurrent neural networks.
\newblock {\em arXiv preprint arXiv:1511.06732\/} .

\bibitem[{Reddy et~al.(2014)Reddy, Lapata, and Steedman}]{reddy2014large}
S.~Reddy, M.~Lapata, and M.~Steedman. 2014.
\newblock Large-scale semantic parsing without question-answer pairs.
\newblock {\em Transactions of the Association for Computational Linguistics
  (TACL)\/} 2(10):377--392.

\bibitem[{Riedel et~al.(2016)Riedel, Bosnjak, and
  Rockt{\"a}schel}]{riedel2016programming}
S.~Riedel, M.~Bosnjak, and T.~Rockt{\"a}schel. 2016.
\newblock Programming with a differentiable forth interpreter.
\newblock {\em CoRR, abs/1605.06640\/} .

\bibitem[{Ross et~al.(2011)Ross, Gordon, and Bagnell}]{ross2011reduction}
S.~Ross, G.~Gordon, and A.~Bagnell. 2011.
\newblock A reduction of imitation learning and structured prediction to
  no-regret online learning.
\newblock In {\em Artificial Intelligence and Statistics (AISTATS)\/}.

\bibitem[{Shen et~al.(2015)Shen, Cheng, He, He, Wu, Sun, and
  Liu}]{shen2015minimum}
S.~Shen, Y.~Cheng, Z.~He, W.~He, H.~Wu, M.~Sun, and Y.~Liu. 2015.
\newblock Minimum risk training for neural machine translation.
\newblock {\em arXiv preprint arXiv:1512.02433\/} .

\bibitem[{Smith and Eisner(2006)}]{smith2006minimum}
D.~A. Smith and J.~Eisner. 2006.
\newblock Minimum risk annealing for training log-linear models.
\newblock In {\em International Conference on Computational Linguistics and
  Association for Computational Linguistics (COLING/ACL)\/}. pages 787--794.

\bibitem[{Sutton et~al.(1999)Sutton, McAllester, Singh, and
  Mansour}]{sutton1999policy}
R.~Sutton, D.~McAllester, S.~Singh, and Y.~Mansour. 1999.
\newblock Policy gradient methods for reinforcement learning with function
  approximation.
\newblock In {\em Advances in Neural Information Processing Systems (NIPS)\/}.

\bibitem[{Venkatraman et~al.(2015)Venkatraman, Hebert, and
  Bagnell}]{venkatraman2015improving}
A.~Venkatraman, M.~Hebert, and J.~A. Bagnell. 2015.
\newblock Improving multi-step prediction of learned time series models.
\newblock In {\em Association for the Advancement of Artificial Intelligence
  (AAAI)\/}. pages 3024--3030.

\bibitem[{Williams(1992)}]{williams1992simple}
R.~J. Williams. 1992.
\newblock Simple statistical gradient-following algorithms for connectionist
  reinforcement learning.
\newblock {\em Machine learning\/} 8(3):229--256.

\bibitem[{Williams and Peng(1991)}]{williams1991function}
R.~J. Williams and J.~Peng. 1991.
\newblock Function optimization using connectionist reinforcement learning
  algorithms.
\newblock {\em Connection Science\/} 3(3):241--268.

\bibitem[{Yin et~al.(2015)Yin, Lu, Li, and Kao}]{yin2015enquirer}
P.~Yin, Z.~Lu, H.~Li, and B.~Kao. 2015.
\newblock Neural enquirer: Learning to query tables.
\newblock {\em arXiv preprint arXiv:1512.00965\/} .

\end{thebibliography}

\appendix

\section{Hyperparameters in \reftab{x-final}}\label{sec:hyperparams}
\newcommand\HYPER[1]{\parbox[c]{1.6cm}{\vspace{.3em}#1\vspace{.3em}}}
{\scriptsize
\begin{tabular}{|c|c|c|c|}
\hline
\textbf{System} & \textsc{Alchemy} & \textsc{Tangrams} & \textsc{Scene} \\
\hline
REINFORCE &
  \HYPER{Sample size $32$\\ Baseline $10^{-2}$\\ $\epsilon = 0.15$\\ embed \textsc{Tokens}} &
  \HYPER{Sample size $32$\\ Baseline $10^{-2}$\\ $\epsilon = 0.15$\\ embed \textsc{Tokens}} &
  \HYPER{Sample size $32$\\ Baseline $10^{-4}$\\ $\epsilon = 0.15$\\ embed \textsc{Tokens}} \\
\hline
BS-MML &
  \HYPER{Beam size $128$\\ embed \textsc{Tokens}} &
  \HYPER{Beam size $128$\\ embed \textsc{Tokens}} &
  \HYPER{Beam size $128$\\ embed \textsc{Tokens}} \\
\hline
\ourmodel &
  \HYPER{$\beta=1$\\ $\epsilon = 0.05$\\ embed \textsc{Tokens}} &
  \HYPER{$\beta=1$\\ $\epsilon = 0.15$\\ embed \textsc{Tokens}} &
  \HYPER{$\beta=0$\\ $\epsilon = 0.15$\\ embed \textsc{Stack}} \\
\hline
\end{tabular}
}

\clearpage
\clearpage
\section{SCONE domains and program tokens}\label{sec:tokens_table}

\newcommand{\bpush}{\textbf{push:} }
\newcommand{\bpop}{\textbf{pop:} }
\newcommand{\bperform}{\textbf{perform:} }
\newcommand{\bhead}[1]{\multicolumn{3}{|p{6.1in}|}{#1}}

{
\small
\renewcommand{\arraystretch}{1.15}
\begin{tabular}{|l|l|l|} \hline
  \textbf{token} & \textbf{type} & \textbf{semantics} \\ \hline
  \bhead{\textbf{Shared across \textsc{Alchemy}, \textsc{Tangrams}, \textsc{Scene}}} \\ \hline
  \C{1}, \C{2}, \C{3}, \dots & constant & \bpush number \\
  \C{-1}, \C{-2}, \C{-3}, \dots & & \\ \hline
  \C{red}, \C{yellow}, \C{green}, & constant & \bpush color \\
  \C{orange}, \C{purple}, \C{brown} & & \\ \hline
  \C{allObjects} & constant & \bpush the list of all objects \\ \hline
  \C{index} & function & \bpop a list $L$ and a number $i$ \\
            & & \bpush the object $L[i]$ (the index starts from 1; negative indices are allowed) \\ \hline
  \C{prevArg$j$} ($j = 1, 2$) & function & \bpop a number $i$ \\
                        & & \bpush the $j$ argument from the $i$th action \\ \hline
  \C{prevAction} & action & \bpop a number $i$ \\
         & & \bperform fetch the $i$th action and execute it
             using the arguments on the stack \\ \hline
    \bhead{\textbf{Additional tokens for the \textsc{Alchemy} domain}

    An \textsc{Alchemy} world contains 7 beakers.
    Each beaker may contain up to 4 units of colored chemical.} \\ \hline
  \C{1/1} & constant & \bpush fraction (used in the \C{drain} action) \\ \hline
  \C{hasColor} & function & \bpop a color $c$ \\
               & & \bpush list of beakers with chemical color $c$ \\ \hline
  \C{drain} & action & \bpop a beaker $b$ and a number or fraction $a$ \\
            & & \bperform remove $a$ units of chemical (or all chemical if $a =$ \C{1/1}) from $b$ \\ \hline
  \C{pour} & action & \bpop two beakers $b_1$ and $b_2$ \\
           & & \bperform transfer all chemical from $b_1$ to $b_2$ \\ \hline
  \C{mix} & action & \bpop a beaker $b$ \\
           & & \bperform turn the color of the chemical in $b$ to \C{brown} \\ \hline
    \bhead{\textbf{Additional tokens for the \textsc{Tangrams} domain}

    A \textsc{Tangrams} world contains a row of tangram pieces with different shapes.
    The shapes are anonymized; a tangram can be referred to by an index or a history reference, but not by shape.} \\ \hline
  \C{swap} & action & \bpop two tangrams $t_1$ and $t_2$ \\
           & & \bperform exchange the positions of $t_1$ and $t_2$ \\ \hline
  \C{remove} & action & \bpop a tangram $t$ \\
             & & \bperform remove $t$ from the stage \\ \hline
  \C{add} & action & \bpop a number $i$ and a previously removed tangram $t$ \\
          & & \bperform insert $t$ to position $i$ \\ \hline
    \bhead{\textbf{Additional tokens for the \textsc{Scene} domain}

    A \textsc{Scene} world is a linear stage with 10 positions.
    Each position may be occupied by a person with a colored shirt
    and optionally a colored hat. There are usually 1-5 people on the stage.} \\ \hline
  \C{noHat} & constant & \bpush pseudo-color (indicating that the person is not wearing a hat) \\ \hline
  \C{hasShirt}, \C{hasHat} & function & \bpop a color $c$ \\
               & & \bpush the list of all people with shirt or hat color $c$ \\ \hline
  \C{hasShirtHat} & function & \bpop two colors $c_1$ and $c_2$ \\
                  & & \bpush the list of all people with shirt color $c_1$ and hat color $c_2$ \\ \hline
  \C{leftOf}, \C{rightOf} & function & \bpop a person $p$ \\
                          & & \bpush the location index left or right of $p$ \\ \hline
  \C{create} & action & \bpop a number $i$ and two colors $c_1$, $c_2$ \\
             & & \bperform add a new person at position $i$ with shirt color $c_1$ and hat color $c_2$ \\ \hline
  \C{move} & action & \bpop a person $p$ and a number $i$ \\
           & & \bperform move $p$ to position $i$ \\ \hline
  \C{swapHats} & action & \bpop two people $p_1$ and $p_2$ \\
               & & \bperform have $p_1$ and $p_2$ exchange their hats \\ \hline
  \C{leave} & action & \bpop a person $p$ \\
            & & \bperform remove $p$ from the stage \\ \hline
\end{tabular}
}

\end{document}